\documentclass[10pt,twocolumn,letterpaper]{article}

\usepackage{cvpr}
\usepackage{times}
\usepackage{epsfig}
\usepackage{graphicx}
\usepackage{amsmath}
\usepackage{amssymb}
\usepackage[table]{xcolor}
\usepackage{float}
\usepackage{textgreek}
\usepackage{enumitem}
\usepackage[utf8]{inputenc}
\usepackage[tableposition=bottom]{caption}
\usepackage{multirow}

\usepackage[pagebackref=true,breaklinks=true,letterpaper=true,colorlinks,bookmarks=false]{hyperref}

\usepackage{indentfirst}

\cvprfinalcopy 

\def\cvprPaperID{****} 

\ifcvprfinal\fi
\begin{document}

\title{ A Thorough Review on Recent Deep Learning Methodologies for Image Captioning}

\author{Ahmed Elhagry\\
{\tt\small 20020054@mbzuai.ac.ae}
\and
Karima Kadaoui\\
{\tt\small 20020048@mbzuai.ac.ae}
}

\maketitle
\begin{abstract}
   Image Captioning is a task that combines computer vision and natural language processing, where it aims to generate descriptive legends for images. It is a two-fold process relying on accurate image understanding and correct language understanding both syntactically and semantically. It is becoming increasingly difficult to keep up with the latest research and findings in the field of image captioning due to the growing amount of knowledge available on the topic. There is not, however, enough coverage of those findings in the available review papers. We perform in this paper a run-through of the current techniques, datasets, benchmarks and evaluation metrics used in image captioning. The current research on the field is mostly focused on deep learning-based methods, where attention mechanisms along with deep reinforcement and adversarial learning appear to be in the forefront of this research topic. In this paper, we review recent methodologies such as UpDown, OSCAR, VIVO, Meta Learning and a model that uses conditional generative adversarial nets. Although the GAN-based model achieves the highest score, UpDown represents an important basis for image captioning and OSCAR and VIVO are more useful as they use novel object captioning.
This review paper serves as a roadmap for researchers to keep up to date with the latest contributions made in the field of image caption generation.
   
\end{abstract}

\section{Introduction}
Image captioning has a variety of applications. From automatic image indexing to assistive technology, there is more and more incentive to build solutions that don't rely on human annotation since it quickly becomes cost inefficient. A lot of work has been done in the field already and some image captioning survey papers are available, but none of them cover some important models that came out between 2018 and 2020. In a field that evolves at this rate, it is imperative to keep track of the latest models and frameworks. That is why we're presenting this paper, in an attempt to cover some of the most relevant papers that have recently been published and compare the performances of their models and describing how they work.

One of the techniques that have a crucial role in image captioning today is the use of attention mechanisms. Ever since transformers \cite{attention} were introduced, a number of different tasks such as machine translation and language modeling saw significant improvements thanks to them. Image captioning is no different: there is an extensive use of top-down visual attention in different models as will be presented in the paper. Another technique that sparked researchers' interest is deep reinforcement learning. It has shown to be particularly good with unusual images (e.g. bed in a forest). The way it's used in image captioning is through optimizing the reward function by maximizing its expected value, which cannot be done using MLE since metrics are not differentiable.

\subsection{Related Work}
As mentioned in the review paper \cite{StudyReview2018}, the authors presented a comprehensive review of the state-of-the-art deep learning-based image captioning techniques by late 2018. The paper gave a taxonomy of the existing techniques, compared the pros and cons, and handled the research topic from different aspects including learning type, architecture, number of captions, language models and feature mapping. They also discussed both the strengths and weaknesses of different datasets and evaluation metrics. According to another review paper \cite{StudyReview2019} published later mid 2019, the authors compared the image captioning methodologies from 2016 to 2019 including newer ones on two datasets: MSCOCO and Flickr30k. An investigation was done on different feature extractors including AlexNet, VGG-16 Net, ResNet, GoogleNet with all the nine Inception models, and DenseNet.  In addition, language models were covered such as LSTM, RNN, CNN, GRU and TPGN. This is while comparing several evaluation metrics including BLEU (1 to 4), CIDEr and METEOR. In \cite{StudyReview2020}, the paper put the spotlight on some of the advancement on the image captioning task until early 2020, where various approaches were discussed including N-cut, color-based segmentation and hybrid engine. It also discussed how model engineering and incorporating more hyper-parameters improve the overall pipeline and result in the best accuracy for such models. In the same year, another study \cite{StudyReview2020_2} covered a review about the literature from 2017 to 2019, where they discussed different datasets and architectures. They stated that the CNN-LSTM models outperformed the CNN-RNN ones, and the most evaluation metric used was BLEU (1 to 4). They also found that the best methods for such a model implementation are encode-decoder and attention mechanism. Furthermore, they mentioned that a combination of both methods can help in improving the results in such a task. Image captioning remains an active research area, and new methodologies keep being published up until this moment. That was one of the main motivations to write this review paper in order to tackle all the recent advances in the past few years including 2020.
\section{Methods}

\subsection{UpDown}
Most common mechanisms that rely on visual attention today are of the top-down kind. They are fed their partially finished caption at each time step to gain context. The issue with these models however is that there is no deliberation as to what regions of an image will receive attention.
This has an effect on the quality of the captions as focusing on salient object regions will provide descriptions that are similar to ones given by humans \cite{objatt}. 

\cite{updown} introduce Up-Down, a model that joins an entirely visual bottom-up mechanism and a task-specific context top-down one. The former gives proposals on image regions that it deems salient, while the latter uses context to compute an attention distribution over them, thus allowing attention to be directed to the important objects in the input image.

\noindent\textbf{Implementation Details}
The bottom-up mechanism employs the Faster R-CNN \cite{faster} object detection model, responsible for recognizing class objects and surrounding them with bounding boxes. 
For pre-training, it is initialized with Resnet-101 \cite{resnet} and trained on the Visual Genome dataset \cite{VG}. The top-down mechanism uses a visual attention LSTM and a language one. The attention LSTM is fed the previous language LSTM outputs, the word generated at time \textit{t-1} and mean-pooled image features to decide which regions should receive attention.
The generated caption up until that point is then used to compute the conditional distribution over potential output words, and the product of all of the conditional distributions gives the distribution over the complete captions. 

\subsection{OSCAR} 
Vision-language pre-training (VLP) is widely used for learning cross-modal representations. It suffers, however, from 2 issues \cite{oscar}: a difficulty in discerning features due to the overlap of their image regions and a lack of alignment caption words and their corresponding image regions. \cite{oscar} remedy to this by using object tags as "anchor points". 
More specifically, they use triples as inputs composed of image region features, object tags and word sequence (caption). 
Doing this helps, because when one channel is incomplete or noisy, the other might complete the information (you can describe an object both through image and language).
It is therefore simple to make the alignments, because the most important elements in the image appear in the matching caption and are also the ones that are expected to receive the most attention.

\begin{figure}
 \centering
 \includegraphics[scale=0.5]{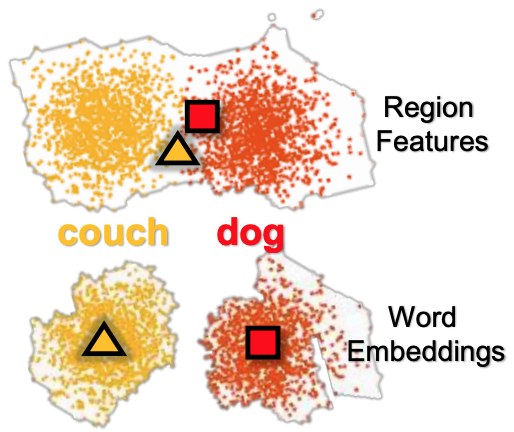}
 \caption{The semantic space used by OSCAR \cite{oscar}. In the example of a dog sitting on a couch, "couch" and "dog" are close in region features since they're roughly in the same area of the image, but they farther apart in word embeddings because of their different meanings.}
 \label{semantic}
\end{figure}

\noindent\textbf{Implementation Details}
OSCAR detects object tags using Faster R-CNN \cite{faster} and presents a 2-view perspective:\\ (1) A dictionary-view with a linguistic semantic space encompassing the tags and caption tokens, and a visual semantic space where image regions lay (Fig. \ref{semantic}). \\
(2) A modality-view that consists of an image modality containing image features and tags, and a language modality with caption tokens. 
The total loss of pretraining is defined by the addition of a masked token loss of predicting masked tokens form the linguistic semantic space and the contrastive loss of predicting if an image-tag sequence is polluted (from replaced tags).


At inference time, the input consists of image regions and tags 
At each time step of the generation, a [MASK] token is appended to the sequence before being replaced by a token from the vocabulary
until the [STOP] token is generated.

\subsection{VIVO}
\begin{figure}
 \centering
 \includegraphics[scale=0.4]{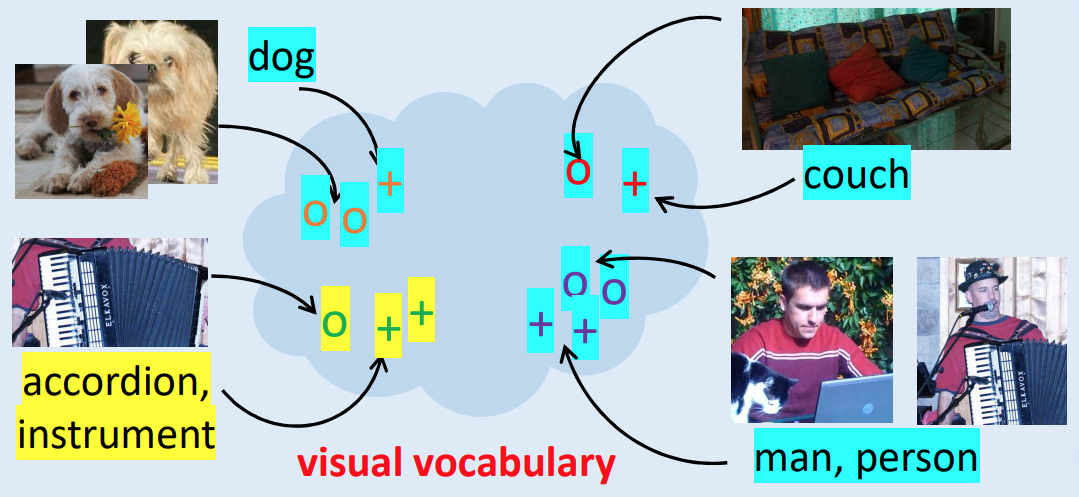}
 \caption{Visual Vocabulary used by VIVO. \cite{vivo} Objects that are similar semantically are closer together. o represents regions and + represents tags. Yellow objects and tags are novel.}
 \label{vivo}
\end{figure}

In the nocaps \cite{nocaps} challenge, the only allowed image-caption dataset is the MS COCO \cite{coco} one, making conventional VLP methods inapplicable \cite{vivo}. For that reason, \cite{vivo} came up with VIVO (VIsual VOcabulary pre-training) 
What it does differently is defining a "visual vocabulary", which is a joint embedding space of tags and image region features, where vectors of semantically close objects (e.g. accordion and instrument) are close to each other (Fig. \ref{vivo}).
After pre-training the vocabulary, the model is fine-tuned with image-caption pairs using the MS COCO dataset \cite{coco}.
The key difference between VIVO and other VLP models is that VIVO is only pre-trained on image-tag pairs and no captions are involved before fine-tuning. This can prove very useful since tags are easier to be generated automatically which allows the use of a huge number of them for no annotation costs.

\noindent \textbf{Implementation Details}
VIVO uses a multi-layer Transformer responsible for aligning tags with their corresponding image region features followed by a linear layer and softmax. During pre-training, image region features are extracted from the input image using Updown's object detector \cite{updown} and fed to the Transformer along with a set of pairs of images and tags. One or more tags are randomly masked and the model makes predictions based on the remaining tags and the image regions. 

In fine-tuning, the model is fed a triplet of image regions, tags and a caption where some of the caption's tokens are randomly masked and the model learns to predict them. A uni-directional attention mask is applied and the parameters are optimized using a cross-entropy loss.

At inference time, image region features are extracted from the input image and tags are detected. A caption is then generated one token at a time in an auto-regressive manner (using the previous tokens as input) until the end token is generated or the caption reached its maximum length.


\subsection{Meta Learning}
One of the drawbacks of reinforcement learning is the reward hacking problem, which, in other words, is overfitting on the reward function which occurs when the agent finds a way to maximize the score without generating captions of a better quality.
When using a CIDEr optimization \cite{scst} for example, common phrases are given less weight and punishment is given to a caption that is too short. As a result, when a short caption is generated, common phrases are added to it to make it longer, ending up with unnatural sentence endings such as "a little girl holding a cat in
a of a." \cite{metaLearning}

\cite{metaLearning} introduce meta learning, which is learning a meta model that is able to optimize and adapt to several different tasks \cite{finn}. In this case, the model simultaneously optimizes the reward function (reinforcement task) and uses supervision from the ground truth (supervision task) by taking gradient steps in both directions. This guarantees the distinctiveness of the captions and their propositional correctness and results in sound human-like sentences. 

\begin{figure}
 \centering
 \includegraphics[scale=0.4]{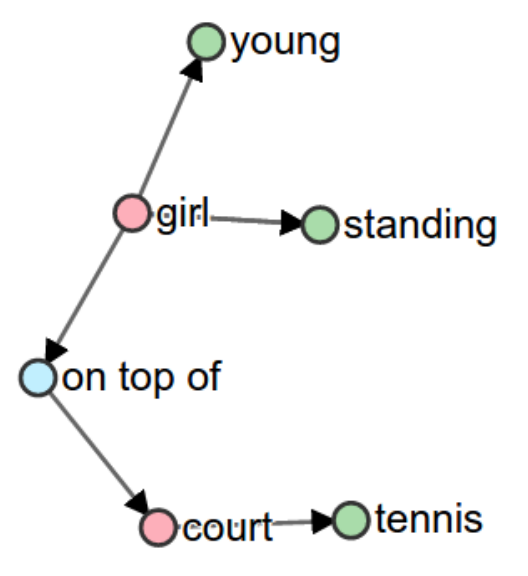}
 \caption{SPICE scene graph for the caption "A young girl standing on top of a tennis court". The objects are marked red, the relations blue and the attributes green. \cite{SPICE}}
 \label{spice}
\end{figure}

Additionally, they import the SPICE \cite{SPICE} metric and add it to the CIDEr \cite{cider} reward term since it performs semantic propositional evaluation using a scene graph. What this means is that an unusual caption ending will be an object-relation pair in the scene graph without a match (Fig. \ref{spice}). Unfortunately, SPICE has a reward hacking issue of its own since it allows duplicate tuples. It is therefore not easy to develop an ideal evaluation metric.

\noindent\textbf{Implementation Details} 
\cite{metaLearning} use the UpDown architecture \cite{updown} as outlined above.
The two tasks that the model needs to optimize are the maximum likelihood estimate task and the reinforcement learning task. In other words, it needs to take 2 gradient steps in order to update the parameter $\theta$. In the first step, the model adapts $\theta$ to the two tasks and calculates their respective losses. Then, $\theta$ updates itself in what is named a "meta update". Doing things this way, the model learns the parameter $\theta$ that optimizes both tasks instead of simply taking a step in between the two gradients when adding up the losses. 

\subsection{Conditional GAN-Based Model} 
To overcome reward hacking, \cite{gan} use discriminator networks to decide whether a generated caption is from a human or a machine.
Since they do not give their model a name, we will call it IC-GAN (Image Captioning GAN) for the sake of practicality. 


\noindent\textbf{Implementation Details}
\cite{gan} experimented with 2 different architectures for the discriminator: one of them using a CNN with a fully connected layer and a sigmoid transformation, and the other an RNN (LSTM) with a fully connected layer and a softmax. They also experiment with an ensemble of 4 CNNs and 4 RNNs. For the generator, a number of different architectures were used but we will focus in the results on the generator that uses the UpDown architecture \cite{updown}. In all cases, the generator and discriminator need to be pre-trained before being alternatively fine-tuned.


\subsection{Evaluation Metrics}
To compare the quality of the generated captions to the ground-truth, a number of evaluation metrics are used. 
The most common used ones being CIDEr, SPICE, BLEU and METEOR. The common metrics across all the covered literature are CIDEr and SPICE, which is why we will be using them.
CIDEr \cite{cider} is an image classification evaluation metric that uses 
term frequency-inverse document frequency (TF-IDF) \cite{TF_IDF} to achieve human consensus.
SPICE \cite{SPICE} is a new semantic concept-based caption assessment metric based on scene-graph, a graph-based semantic representation (Fig. \ref{spice}) \cite{SPICE_2}\cite{sPICE_3}. 
\subsection{Benchmarks}
\cite{nocaps} develop a benchmark called nocaps, which, in addition to the image captioning dataset Microsoft COCO Captions \cite{coco}, makes use of the Open Images object detection dataset \cite{openimages} to introduce novel objects not seen in the former. The nocaps benchmark is made up of 166,100 captions that describe 15,100 images from the OpenImages validation and test sets. 
OSCAR \cite{oscar} and VIVO \cite{vivo} methods are evaluated on the nocaps validation set \cite{nocaps}. Karpathy splits \cite{karpathy} are used in the evaluation of the meta learning model \cite{metaLearning} and the IC-GAN one \cite{gan}. 
Finally, UpDown \cite{updown} is evaluated on both benchmarks.

\section{Results}

\begin{table}[t]
  \centering
  \begin{tabular}{  p{5cm}||p{1cm} p{1cm}}
 \hline
\multicolumn{1}{c||}{Method}  & CIDEr & SPICE\\
 \hline
 Resnet Baseline & 111.1 & 20.2\\
 UpDown &  120.1 & 21.4\\
 \hline
 MLE Maximization & 110.2 & 20.3\\
 *RL Maximization & 120.4 & 21.3\\
 *MLE + RL Maximization & 119.3 & 21.2\\
 *Meta Learning & 121.0 &  21.7\\
 \hline
 IC-GAN (Updown/CNN-GAN) & 123.2 & 22.1\\
 IC-GAN (Updown/RNN-GAN) & 122.2 & 22.0 \\
 IC-GAN (Updown/ensemble) & \textbf{125.9} &\textbf{22.3} \\

 \hline
 \end{tabular}
  \caption{Results of the overall performance on MS COCO Karpathy test split \cite{metaLearning} \cite{gan}. Methods with a * symbol use reinforcement learning with CIDEr optimization.}
  \label{tab1}
\end{table}  

\begin{table*}[t]
  \centering
  \arraybackslash
  \begin{tabular}{m{3cm}||p{1cm} p{1cm}|p{1cm} p{1cm}| p{1cm} p{1cm}| p{1cm} p{1cm}}
\multirow{2}{*}{Method} & \multicolumn{2}{c|}{in-domain} & \multicolumn{2}{c|}{near-domain} & \multicolumn{2}{c|}{out-of-domain} & 
\multicolumn{2}{c}{overall}\\
   &  CIDEr & SPICE &  CIDEr & SPICE &  CIDEr & SPICE &  CIDEr & SPICE\\
 \hline
  \multicolumn{9}{c}{Validation Set} \\
  \hline
 UpDown (2019)  & 78.1 & 11.6 & 57.7 & 10.3 & 31.3 & 8.3 & 55.3 & 10.1 \\
 UpDown + CBS  & 80.0 & 12.0 & 73.6 & 11.3 & 66.4 & 9.7 & 73.1 & 11.1\\
UpDown + ELMo + CBS & 79.3 & 12.4 & 73.8 & 11.4 & 71.7 & 9.9 & 74.3 & 11.2\\
 \hline
 OSCAR (2020)  & 79.6 & 12.3 & 66.1 & 11.5 & 45.3 & 9.7 & 63.8 & 11.2\\
 OSCAR + CBS & 80.0 & 12.1 & 80.4 & 12.2 & 75.3 & 10.6 & 79.3 & 11.9\\
 OSCAR+SCST+CBS & 83.4 & 12.0 & 81.6 & 12.0 & 77.6 & 10.6 & 81.1 & 11.7\\
 \hline
 VIVO (2020) & 88.8 & 12.9 & 83.2 & 12.6 & 71.1 & 10.6 & 81.5 & 12.2 \\
 VIVO + CBS & 90.4 & 13.0 & 84.9 & 12.5 & 83.0 & 10.7 & 85.3 & 12.2\\
 VIVO+SCST+CBS & \textbf{92.2} & 12.9 & \textbf{87.8} & 12.6 & 87.5 & 11.5 & \textbf{88.3} & 12.4\\
 \hline
 Human & 84.4 & \textbf{14.3} & 85.0 & \textbf{14.3} & \textbf{95.7} & \textbf{14.0} & 87.1 & \textbf{14.2}\\
 \hline
 \end{tabular}
  \caption{Evaluation on nocaps validation set \cite{vivo}}
  \label{tab2}
\end{table*}

\subsection{MS COCO Karpathy Splits Benchmark}
\cite{updown} initially ran experiments on both an ablated Resnet \cite{resnet} baseline model and the UpDown model to measure the impact of the bottom-up attention. Since the meta learning model \cite{metaLearning} and IC-GAN \cite{gan} use the UpDown architecture themselves, we have decided to include the Resnet baseline to the comparison (Table \ref{tab1}) to get an idea about the individual effect of the bottom-up + top-down approach. 

We notice that UpDown shows an important gain in performance going from 111.1 to 120.1 and from 20.2 to 20.4 in the CIDEr and SPICE metrics respectively. This represents a relative improvement of 8\% in CIDEr and 3\% in SPICE. Therefore, adding bottom-up attention has an important positive impact on image captioning.

Concerning the experiments in \cite{metaLearning}, we observe that the model that uses meta learning receives a CIDEr score of 121.0 and a SPICE score of 21.7. It is the most performant on both evaluation metrics compared to maximizing using the maximum likelihood estimate, reinforcement learning and the MLE+RL maximization (which relies on simply adding up the gradients from the supervision and reinforcement tasks). It also has a slight improvement on the UpDown model without the use of meta learning.

IC-GAN on the other hand, shows the highest performance with a significant improvement on all 3 models. The relative improvements compared to the UpDown model range between 1.7\% (RNN-GAN) and  4.6\% (ensemble). Additionally, comparing with conventional reinforcement learning approaches, the proposed adversarial learning method boosts the performance from 8.9\% to 16.3\% \cite{gan}.

It is important to note that although using a CNN-GAN slighlty improves the score compared to using an RNN-GAN, the latter can save up to 30\% of training time compared to the former \cite{gan}.

Evaluation scores aside, IC-GAN is able to generate human-like captions and avoids mistakes commonly made by tradition reinforcement learning methods, such as duplicated words and logical errors like "a group of people standing on top of a clock" \cite{gan}. 

\subsection{nocaps Benchmark}
The OSCAR model \cite{oscar} is characterized by being highly efficient when it comes to parameters due to the anchor points making the semantic alignments learning easier. When used on its own, it outperfoms the UpDown model on all in-domain, near-domain and out-of-domain subsets.
By adding Constrained Beam Search \cite{cbs} and Self Critical Sequential Training (SCST) \cite{scst} the performance improves tremendously, particularly in the out-of-domain subset going from 45.3 to 77.6 in CIDEr and from 9.7 to 10.6 in SPICE.

OSCAR does not score however, as high as VIVO as shown in Table \ref{tab2}. In the in-domain case, VIVO on its own outperforms the combinations OSCAR+SCST+CBS and UpDown+ELMo+CBS \cite{elmo} by a difference ranging from 5.4 to 9.5 in CIDEr and from 0.5 to 0.9 on SPICE. 

The VIVO+SCST+CBS version shows the highest performance with CIDEr scores that even surpass the human ones across in-domain and near-domain. The out-of-domain subset results still show great scores that surpass all other models.

\section{Discussion}
It seems that the current research on image captioning is heavily focused on deep-learning techniques, and for a good reason. Image captioning is a very complex task that combines both computer vision and natural language processing and it therefore needs powerful techniques that could handle that level of complexity. Attention mechanisms along with deep reinforcement and adversarial learning, appear to be actively-researched methods for this task, as showcased in this paper. Faster R-CNN is a popular network choice, along with the LSTM. In particular, the UpDown model seems to be used as a basis for multiple papers that were published between 2018 and 2020 which gives it a key role and notable impact on the advances in the field. Novel object captioning also seems to be gathering a lot of interest after proving its usefulness.

Although the VIVO and OSCAR models don't show scores as high as the ones that use meta learning and adversarial learning, they are superior in terms of their usability "in the wild", since the MS COCO dataset \cite{coco} on which all of these models are trained contains only a small part of the objects that we run into in real life. We should therefore not take evaluation scores at face value, especially after demonstrating the reward hacking problem in reinforcement learning. In the future, additional efforts should be put into making more robust reward functions and into more research on novel object captioning, especially replacing human-annotated object detection datasets with fully machine-generated tags.

\section{Conclusions}
Image captioning is an active research topic that creates the space for competition among researchers. It is noticed that the existing review papers do not cover some of the important recent advances, although there are new methodologies with outperforming performance. The outgoing research on image captioning is focused on deep learning-based methods, where attention mechanisms exist along with deep reinforcement and adversarial learning. Our paper discusses the recent methods and their implementations. State-of-the-art techniques include Updown, OSCAR, VIVO, Meta Learning and a GAN-based model. The GAN-based model is the most performant, UpDown has the most impact and OSCAR and VIVO are the more useful. We hope this review will provide the community with a complementary guideline along with the existing review papers for further pursued research in the image captioning research topic.
{\small
\bibliographystyle{ieee}
\bibliography{egbib}
}

\end{document}